\documentclass[10pt,twocolumn,letterpaper]{article}
\usepackage[pagenumbers]{cvpr}      

\usepackage[utf8]{inputenc} 
\usepackage[T1]{fontenc}    
\usepackage{url}            
\usepackage{booktabs}       
\usepackage{amsfonts}       
\usepackage{nicefrac}       
\usepackage{microtype}      

\usepackage{graphicx}
\usepackage{amsmath}
\usepackage{amssymb}

\usepackage{diagbox}
\usepackage{multicol}
\usepackage{enumerate}
\usepackage{times}
\usepackage{epsfig}
\usepackage{threeparttable}
\usepackage{enumitem}
\usepackage{multirow}
\usepackage{color}
\usepackage{array}
\usepackage{setspace}
\usepackage{makecell}
\usepackage{indentfirst}
\usepackage{pifont}
\usepackage{verbatimbox}

\newcommand{\cmark}{\checkmark} 
\newcommand{\xmark}{\ding{55}}   

\definecolor{cvprblue}{rgb}{0.21,0.49,0.74}
\usepackage[pagebackref,breaklinks,colorlinks,allcolors=cvprblue]{hyperref}

\def\paperID{229} 
\def\confName{CVPR}
\def\confYear{2026}

\title{4D-VGGT: A General Foundation Model with SpatioTemporal Awareness \\ for Dynamic Scene Geometry Estimation}

\author{%
	Haonan Wang$\mathbf{^{1}}$, Hanyu Zhou$\mathbf{^{2}}$\footnotemark[1] , Haoyue Liu$\mathbf{^{1}}$, Luxin Yan$\mathbf{^{1}}$\\
	$\mathbf{^{1}}$National Key Lab of Multispectral Information Intelligent Processing Technology,\\
	School of Artificial Intelligence and Automation, Huazhong University of Science and Technology\\
	$\mathbf{^{2}}$School of Computing, National University of Singapore \\
	\texttt{\{whn\_aurora,yanluxin\}@hust.edu.cn, hy.zhou@nus.edu.sg} \\
}

\usepackage[sort&compress]{natbib}
\begin{document}
	\twocolumn[{
		\maketitle
		\setlength{\abovecaptionskip}{6pt}
		\setlength{\belowcaptionskip}{10pt}
		\centering
		\includegraphics[scale=0.38]{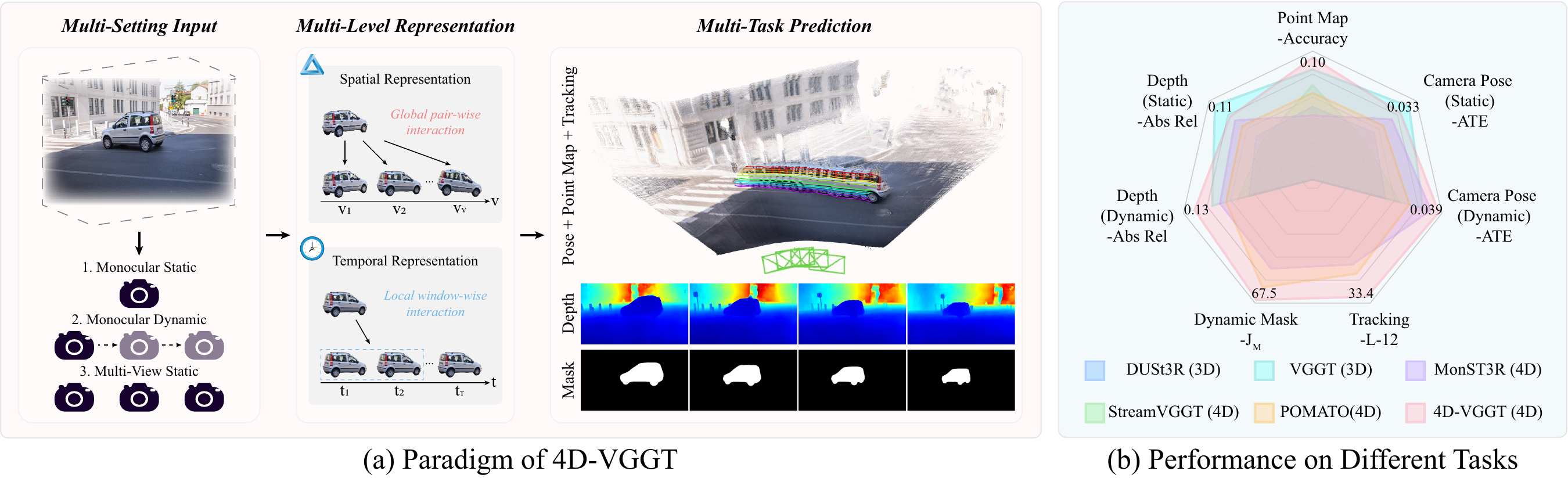}
		\captionof{figure}{\textbf{Illustration of our model paradigm and performance.} (a) Our 4D-VGGT accommodates various camera settings and adopts a divide-and-conquer spatiotemporal representation approach for different geometry tasks in dynamic scenes. (b) Performance comparison of visual geometry models. Our 4D-VGGT achieves consistently superior performance across various geometry tasks.
		}
		\label{fig1}
	}]
	\renewcommand{\thefootnote}{*}   
	\footnotetext[1]{Corresponding author.}
	
	\begin{abstract}
		
		We investigate a challenging task of dynamic scene geometry estimation, which requires representing both spatial and temporal features. Typically, existing methods align the two features into a unified latent space to model scene geometry. However, this unified paradigm suffers from potential mismatched representation due to the heterogeneous nature between spatial and temporal features. In this work, we propose \textbf{4D-VGGT}, a general foundation model with divide-and-conquer spatiotemporal representation for dynamic scene geometry. Our model is divided into three aspects: 1) \textbf{Multi-setting input.} We design an adaptive visual grid that supports input sequences with arbitrary numbers of views and time steps. 2) \textbf{Multi-level representation.} We propose a cross-view global fusion for spatial representation and a cross-time local fusion for temporal representation. 3) \textbf{Multi-task prediction.} We append multiple task-specific heads to spatiotemporal representations, enabling a comprehensive visual geometry estimation for dynamic scenes. Under this unified framework, these components enhance the feature discriminability and application universality of our model for dynamic scenes. In addition, we integrate multiple geometry datasets to train our model and conduct extensive experiments to verify the effectiveness of our method across various tasks on multiple dynamic scene geometry benchmarks.
		
	\end{abstract}
	
	\section{Introduction}
	\label{sec:intro}
	
	
	Foundation models aim to learn general-purpose representations from massive data and are widely applied in diverse vision tasks, such as pose \cite{teed2023deep,chen2024leap}, depth \cite{wang2023neural,yang2024depth,chen2025video,hu2025depthcrafter}, and point trajectories \cite{doersch2023tapir,xiao2024spatialtracker,karaev2024cotracker3}. Recently, some researchers attempt to leverage the strong representation capacity of foundation models to learn the spatial features for scene geometry estimation through multi-task learning, \emph{e.g.}, VGGT \cite{wang2025vggt} and $\pi$3 \cite{wang2025pi}. Despite achieving great success in static scenes, these 3D methods still face significant challenges in dynamic scenes. Different from static scenes, dynamic scene geometry estimation requires the representation of both spatial and temporal features, which exceeds the representation scope of these 3D models. In this work, our goal is to learn the spatiotemporal representations for dynamic scene geometry.
	
	
	Existing methods typically embed temporal cues into spatial features, thus representing the spatiotemporal features within a unified latent space. For instance, Zhang \emph{et al}. \cite{zhang2024monst3r} extend the 3D model DUSt3R \cite{wang2024dust3r} by estimating pair-wise point maps from spatial features and performing temporal consistency alignment. Zhuo \emph{et al}. \cite{zhuo2025streaming} augment VGGT \cite{wang2025vggt} with a temporal causal attention after spatial attention. However, these methods may suffer from the potential mismatched representation due to the heterogeneous nature between spatial and temporal features. This issue further causes these 4D models to produce unstable and unreliable geometry knowledge in dynamic scenes. Therefore, designing a divide-and-conquer spatiotemporal representation is crucial for dynamic scene geometry estimation.
	
	
	The design of our method is based on two key insights: 
	1) \emph{Factorized spatiotemporal representation enhance discriminability.} The spatiotemporal features in dynamic scenes exhibit significant heterogeneity: the spatial dimension follows structural consistency across different views, while the temporal dimension adheres to motion continuity across adjacent time steps. Leveraging this property is beneficial for representing the spatiotemporal characteristics of dynamic scenes.
	2) \emph{Adaptive visual representation improves universality.} 
	Considering that conventional 3D geometry foundation models are capable of handling arbitrary views, we further extend 4D models to support inputs from arbitrary views and arbitrary time steps. This alteration improves the model universality in real-world applications.
	
	
	In this work, we propose \textbf{4D-VGGT}, a general spatiotemporal foundation model for dynamic scene geometry estimation. As illustrated in Fig. \ref{fig1}, our model consists of three main components: 1) \textbf{Multi-Setting Input.} We design an adaptive visual grid that enables our model to accommodate visual features from diverse camera configurations through attention masks.
	2) \textbf{Multi-Level Representation.} We propose a cross-view global fusion module to learn the spatial representation between various views, and a cross-time local fusion to model the temporal representation along continuous time steps.
	3) \textbf{Multi-Task Prediction.} We construct multiple task-specific heads and perform joint multi-task optimization to learn the corresponding spatiotemporal features for scene geometry. Under our unified framework, these components enable our model to utilize a shared spatiotemporal representation scheme to support diverse input configurations and accomplish a variety of visual tasks. Additionally, we also integrate diverse datasets to train our model for more effective dynamic scene geometry estimation. Our main contributions are summarized as follows:
	\begin{itemize}
		\item We present 4D-VGGT, a general spatiotemporal foundation model for dynamic scene geometry estimation. Our model supports multi-setting input, multi-level representation, and multi-task prediction, providing a new modeling paradigm for dynamic scenes.
		\item We observe that visual features extracted from various camera settings share similar distributions. This motivates the design of adaptive visual grid to enable our model to handle vision inputs with arbitrary views and time steps.
		\item We design cross-view global fusion and cross-time local fusion for feature discriminability. The former learns spatial representations across views, while the latter models temporal representations across adjacent time steps.
		\item We integrate diverse geometry datasets to train our model. Extensive experiments demonstrate the superiority of our method across various geometry tasks in dynamic scenes.
	\end{itemize}
	
	
	\begin{figure*}
		\centering
		\setlength{\abovecaptionskip}{0.3cm} 
		\setlength{\belowcaptionskip}{-0.3cm} 
		\includegraphics[scale=0.32]{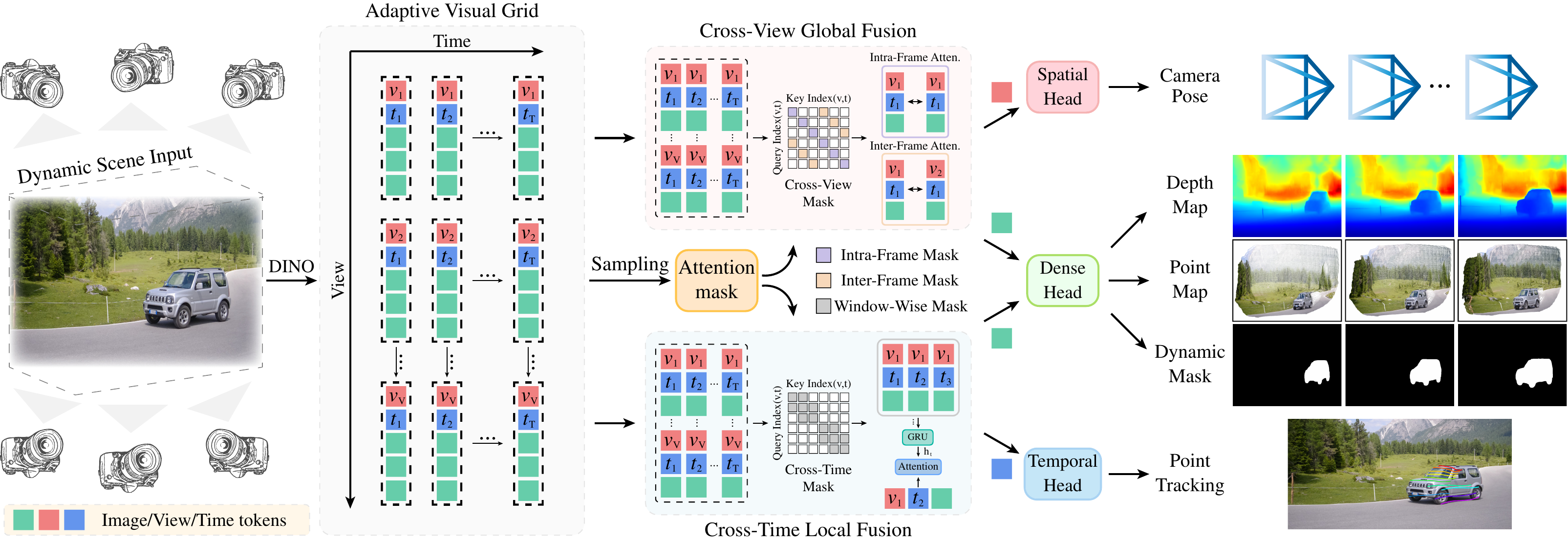}
		\caption{\textbf{Framework of our 4D-VGGT.} Our 4D-VGGT consists of three parts: 1) Multi-setting vision input. Encode  input sequence with DINO and construct adaptive visual grid. 2) Multi-level feature representation. Respectively capture spatial and temporal features in a divide-and-conquer manner. 3) Multi-task geometry prediction. Obtain results for multiple geometry tasks by specific prediction heads.}
		\label{fig2}
	\end{figure*}
	
	\section{Related Work}
	\label{sec:related}
	
	\subsection{3D Geometry Estimation}
	The key to 3D geometry estimation lies in obtaining the spatial representation of static scenes. 3D methods can be broadly divided into optimization-based and learning-based approaches. Optimization-based methods, such as SfM \cite{agarwal2011building,frahm2010building,liu2024robust,schonberger2016structure,wu2013towards}, MVS \cite{furukawa2015multi,galliani2015massively,schonberger2016pixelwise,wang2023adaptive}, and visual SLAM \cite{teed2021droid,mur2015orb,mur2017orb,newcombe2011dtam,campos2021orb}, estimate static scene geometry by optimizing objective functions. However, their assumption of scene rigidity fails in dynamic environments with moving objects. Learning-based methods leverage deep networks to learn 3D priors of static scenes to estimate geometry properties such as pose, depth, and point trajectories. These methods either combine deep networks with traditional optimization \cite{sarlin2020superglue,sun2021loftr,wei2020deepsfm,wang2024vggsfm,zhou2025std}, or use feedforward neural networks with point map representation \cite{wang2024dust3r,leroy2024grounding,wang20243d,wang2025vggt,wang2025pi,yang2025fast3r,tang2025mv,zhang2025flare,elflein2025light3r,liu2025slam3r,deng2025vggt} to estimate scene geometry in a data-driven manner. The lack of temporal modeling limits their performance under dynamic conditions. Therefore, our method additionally introduces a temporal representation module to capture the varying dynamics patterns.
	
	\subsection{4D Geometry Estimation}
	4D geometry estimation requires effective spatiotemporal representation due to the temporally-varying dynamics patterns in dynamic scenes. Existing methods directly embed temporal cues into spatial features for spatiotemporal representation. Some of these methods \cite{zhang2024monst3r,jin2024stereo4d,han2025d,mai2025can,sucar2025dynamic,jiang2025geo4d,feng2025st4rtrack,zhang2025pomato,chen2025easi3r} obtain the local pair-wise point maps based on DUSt3R \cite{wang2024dust3r} and then perform global alignment using the temporal consistency constraint. The others utilize additional temporal modules such as memory-based framework \cite{wang2025continuous,cabon2025must3r,wu2025point3r,cai2025mem4d,zhuo2025streaming} or dynamic-aware module \cite{wang2025c4d} to enhance spatial representation module. However, these direct paradigms are inefficient for dynamic scene geometry estimation since they lead to mismatched representations due to the heterogeneous nature between spatial and temporal features. In this work, we propose a divide-and-conquer strategy and explore two masked-attention modules to separately representing spatial and temporal features.
	
	\subsection{Multi-Task Foundation Models}
	Recent multi-task foundation models aim to unify multiple perception or geometry tasks within a single architecture, typically by sharing a common encoder–decoder backbone across tasks such as depth estimation, semantic segmentation, and normal prediction \cite{zamir2018taskonomy,xiao2024florence,srivastava2024omnivec2,ye2023joint,zhou2025uni4d,zhou2025llava}. While these unified frameworks effectively exploit shared visual representations, they often struggle to maintain task-specific distinctions, leading to suboptimal performance when tasks exhibit conflicting objectives. In this work, we follow a shared-but-distinct strategy and introduce various attention masks to guide the shared transformer-based spatiotemporal fusion module to perform distinct forms of attention computation for different tasks. This strategy can better adapt to different task requirements while keeping the framework universal.
	
	\section{Our 4D-VGGT}
	\label{sec:method}
	\subsection{Overall Framework}
	We propose 4D-VGGT, a general spatiotemporal foundation model for dynamic scene geometry estimation. As shown in Fig. \ref{fig2}, the whole framework is divided into three parts: multi-setting input for adapting and encoding input sequences with arbitrary numbers of views and time steps, multi-level representation for divide-and-conquer spatiotemporal modeling, and multi-task prediction for dynamic scene geometry. In summary, our framework maps each frame $I_{vt}$ to its corresponding geometry attributes, including camera parameter $g_{vt}\in\mathbb{R}^9$, depth map $D_{vt}\in\mathbb{R}^{H\times W}$, dynamic mask $M_{vt}\in\mathbb{R}^{H\times W}$, point map $P_{vt}\in\mathbb{R}^{3\times H\times W}$ and tracking results $T_{vt}$ including 2D trajectory $T_{vt}^{2D}\in\mathbb{R}^{2\times N}$ and 3D trajectory $T_{vt}^{3D}\in\mathbb{R}^{3\times N}$, which is formulated as:
	\begin{equation}\small
		\setlength\abovedisplayskip{3pt}
		\setlength\belowdisplayskip{3pt}
		\mathit{f}((I_{vt})_{v,t=1}^{V,T})=(g_{vt},D_{vt},M_{vt},T_{vt})_{v,t=1}^{V,T}.
	\end{equation}
	
	The entire framework is end-to-end learnable, highly versatile for diverse input sequences and output tasks, and consistently effective as demonstrated by experimental results.
	
	\subsection{Multi-Setting Vision Input}
	\noindent
	\textbf{Various Camera Settings.} Dynamic scenes can be captured under diverse camera configurations, including monocular-static, monocular-dynamic, and multi-view static setups. These configurations create significant variations in input distributions due to differences in appearance and motion patterns, making it difficult for a unified model to perform consistently across different settings. Therefore, it is crucial to explore an adaptive and invariant feature representation to effectively handle various camera settings.

\begin{figure}
		\centering
		\setlength{\abovecaptionskip}{0.3cm} 
		\setlength{\belowcaptionskip}{-0.5cm} 
		\includegraphics[scale=0.32]{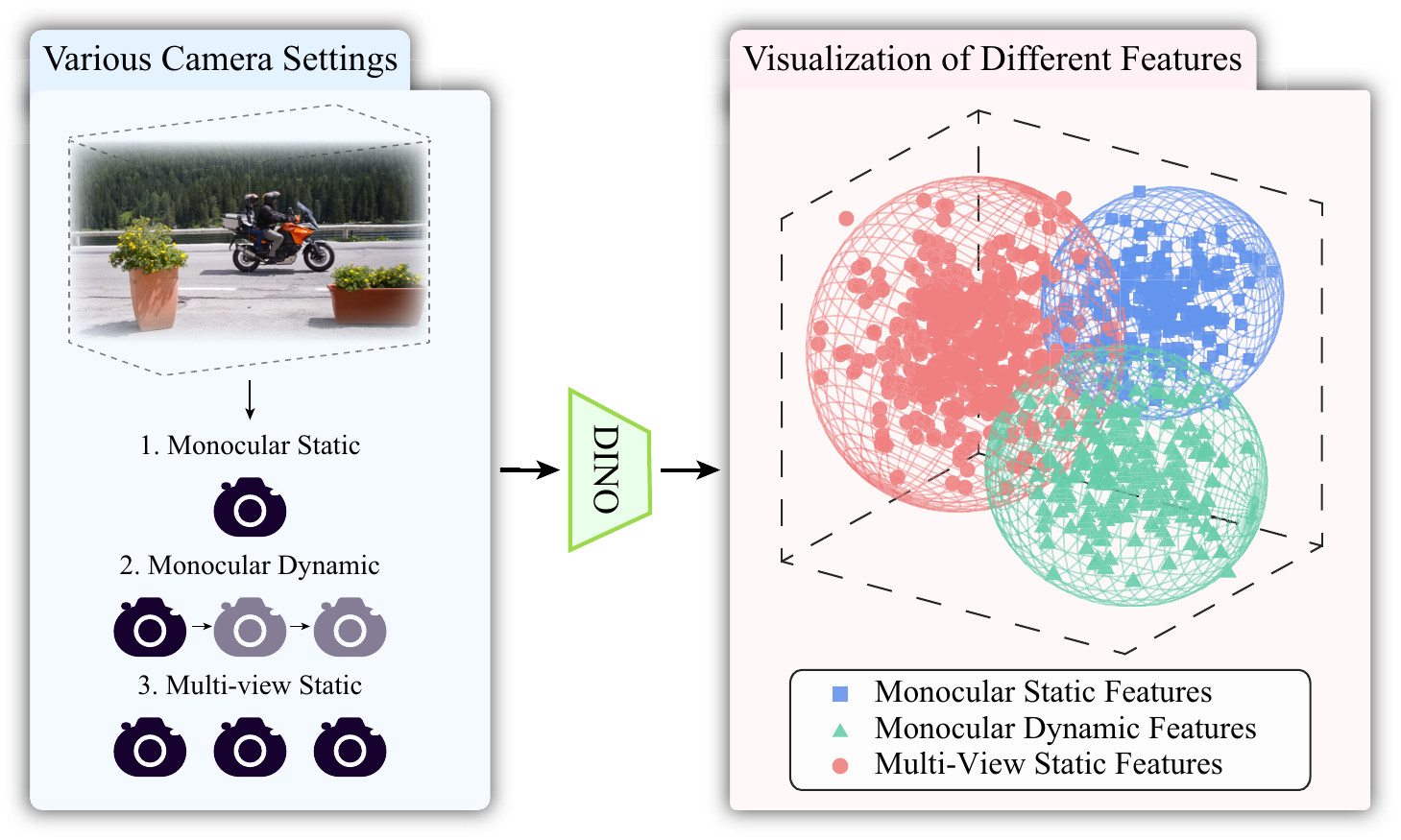}
		\caption{\textbf{Feature distribution of different camera settings.} We use pretrained DINO to analyze the distribution of different camera settings. The similar feature distribution across different settings motivates us to design the adaptive visual grid that accommodates input sequences with arbitrary numbers of views and time steps.}
		\label{fig3}
	\end{figure}

	\noindent
	\textbf{Adaptive Visual Grid.} To explore the possibility of unifying inputs from different camera settings at the feature level, we analyze their feature distributions. We utilize the pretrained DINOv2 model \cite{oquab2023dinov2} to encode image sequences captured under three camera settings from the same scene and apply UMAP \cite{mcinnes2018umap} to visualize the feature distributions. As shown in Fig. \ref{fig3}, features from different views and time steps within the same setting, as well as features across different settings, cluster closely together. This demonstrates spatiotemporal consistency within a setting and cross-setting consistency after DINO encoding. The former motivates our cross-view and cross-time fusion modules (Sec. \ref{sec3.3}) for capturing spatiotemporal consistency, while the latter inspires the design of our adaptive visual grid to accommodate input sequences with arbitrary numbers of views and time steps.
	
	For an input image sequence $I_{vt}$, we first use DINOv2 \cite{oquab2023dinov2} to encode frames into image tokens $t^{I}$. In addition, to identify their view and time information, we add a view token $t^{V}$ and a time token $t^T$ corresponding to each image token. In the adaptive visual grid, we construct a two-dimensional feature grid with view and time coordinates:
	\begin{equation}\small
		\setlength\abovedisplayskip{3pt}
		\setlength\belowdisplayskip{3pt}
		t_{vt}=[t^I_{vt}, t^V_{vt}, t^T_{vt}], t\in[1,T], v\in[1,V].
	\end{equation}
	
	To further improve generalization, we adopt a random sampling strategy during the training stage. We randomly sample some tokens from the visual grid for each training step, simulating various configurations of input sequences. In summary, the adaptive visual grid enables our 4D-VGGT to handle vision inputs with arbitrary views and time steps.
	
	\subsection{Multi-Level Feature Representation}
	\label{sec3.3}
\begin{figure}[t]
		\centering
		\setlength{\abovecaptionskip}{0.3cm} 
		\setlength{\belowcaptionskip}{-0.5cm} 
		\includegraphics[scale=0.29]{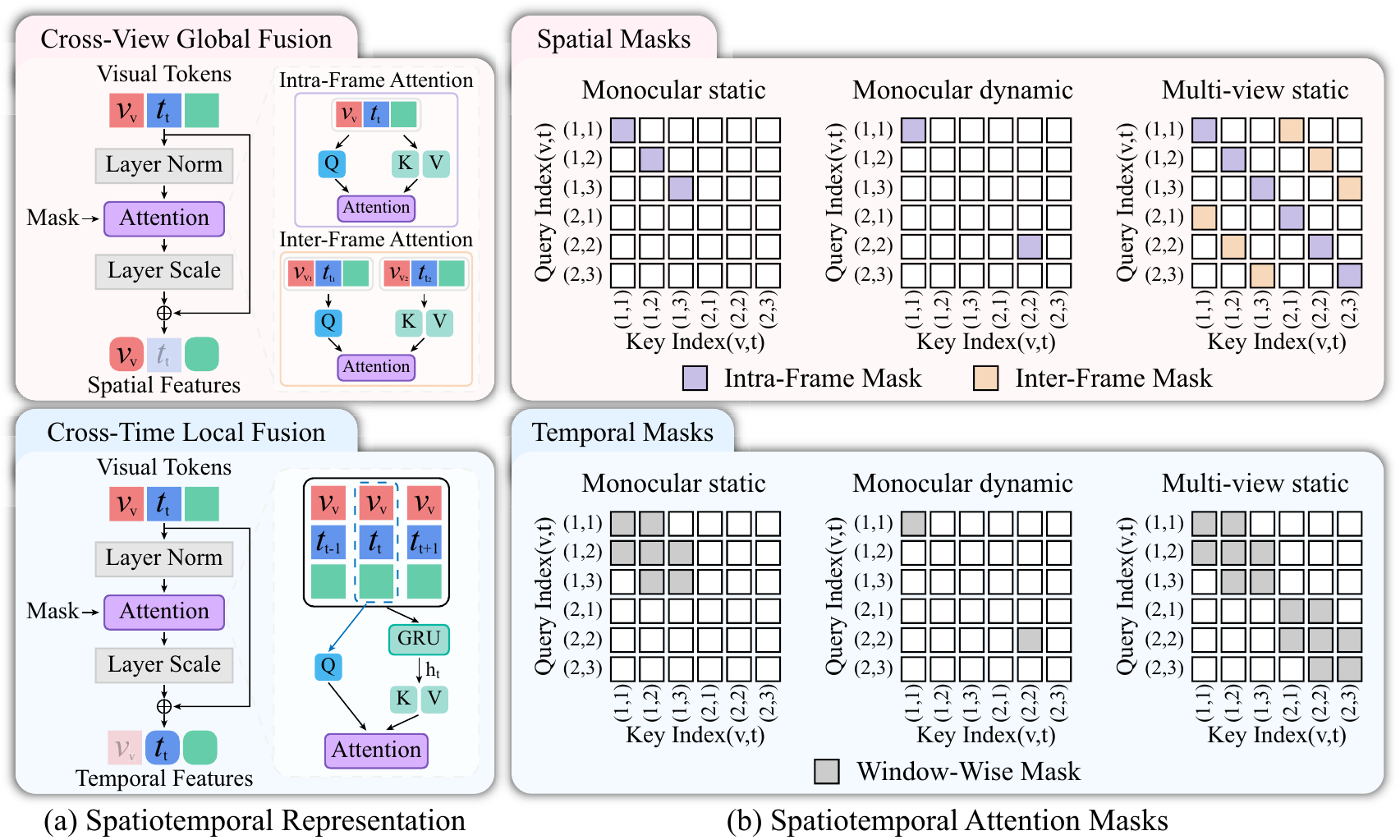}
		\caption{\textbf{Illustration of spatiotemporal representation modules and attention masks.} The visual tokens are input into the masked attention module, which performs different attention calculations guided by the attention masks. The spatial masks enable interactions among all tokens from different views within the same time step, while the temporal masks restrict interactions to tokens from the same view within a fixed temporal window.}
		\label{fig4}
	\end{figure}

	\noindent
	\textbf{Spatiotemporal Representation with Attention.} The spatiotemporal representation of input sequences under different camera settings requires distinct attention interactions. We adopt masked attention that uses specific masks to guide corresponding operations. As illustrated in Fig. \ref{fig4} (b), six attention masks are designed for two modules under three typical camera settings. Specifically, in the cross-view global fusion module, the mask allows interactions among all tokens from different views within the same time step, while in the cross-time local fusion module, it restricts interactions to tokens in a temporal window within the same view.
	
	\noindent
	\textbf{Cross-View Global Fusion.} As shown in Fig. \ref{fig4} (a), guided by the attention masks, the cross-view global fusion module performs two types of attention within each time step: intra-frame attention, which focuses on key areas within a single frame, and inter-frame attention, which captures spatial correlations across frames. After applying these two attention modules $L$ times, we obtain the output spatial feature $F^{S}$. Note that time tokens are used only to identify each token’s time step and do not participate in attention computation. This module is formulated as:
	\begin{equation}\small
		\setlength\abovedisplayskip{3pt}
		\setlength\belowdisplayskip{3pt}
		\begin{aligned}
			\mathit{f}_{CV}([t^I,t^V]) = F^S.
			\label{eq:attention}
		\end{aligned}
	\end{equation}
	
	\begin{table}[t]\scriptsize
		\setlength{\abovecaptionskip}{0.1cm} 
		\setlength{\belowcaptionskip}{0.3cm} 
		\setlength{\tabcolsep}{3.2pt}
		\centering
		\renewcommand{\arraystretch}{0.9}
		\caption{Overview of our training datasets.}
		\addvbuffer[0pt -12pt]{
			\begin{tabular}{cccc}
				\toprule
				Training Stage & Task & Samples & Data Source \\		
				\midrule
				\multirow{13}{*}{\shortstack{Stage-1\\Per-Task Optim.}} & \multirow{2}{*}{\shortstack{Camera\\Pose}} & \multirow{2}{*}{7.1M} & \tiny Co3Dv2 \cite{reizenstein2021common}, WildRGB-D \cite{xia2024rgbd}, ScanNet \cite{dai2017scannet}, \\
				& & & \tiny Hypersim \cite{roberts2021hypersim}, Habitat \cite{ramakrishnan2021habitat}, Replica \cite{straub2019replica} \\
				\cmidrule(lr){2-4}
				& \multirow{2}{*}{Depth} & \multirow{2}{*}{15.6M} &\tiny BlendedMVS \cite{yao2020blendedmvs}, DL3DV \cite{ling2024dl3dv}, MegaDepth \cite{li2018megadepth}, \\
				& & &\tiny Kubric \cite{greff2022kubric}, MVS-Synth \cite{huang2018deepmvs}, Virtual KITTI 2 \cite{cabon2020virtual} \\
				\cmidrule(lr){2-4}
				& \multirow{2}{*}{\shortstack{Dynamic\\Mask}} & \multirow{2}{*}{5.8M} &\tiny Kubric \cite{greff2022kubric}, HOI4D \cite{liu2022hoi4d}, \\
				& & &\tiny Dynamic-Replica \cite{karaev2023dynamicstereo}\\
				\cmidrule(lr){2-4}
				& \multirow{2}{*}{\shortstack{Point\\Map}} & \multirow{2}{*}{1.6M} &\tiny Co3Dv2 \cite{reizenstein2021common}, MegaDepth \cite{li2018megadepth}, \\
				& & &\tiny WildRGB-D \cite{xia2024rgbd}, ScanNet \cite{dai2017scannet}\\
				\cmidrule(lr){2-4}
				& Tracking & 1.2M & \tiny TAP-Vid \cite{doersch2022tap} \\
				\midrule
				\multirow{2}{*}{\shortstack{Stage-2\\Multi-Task Optim.}} & \multirow{2}{*}{Multi-task} & \multirow{2}{*}{8.2M} &\tiny TartanAir \cite{wang2020tartanair}, Spring \cite{mehl2023spring},\\
				& & &\tiny Omniworld \cite{zhou2025omniworld}, SpatialVID \cite{wang2025spatialvid}\\
				\bottomrule
			\end{tabular}
		}
		\label{tab1:datasets}
	\end{table}
	
	\noindent
	\textbf{Cross-Time Local Fusion.} As shown in Fig. \ref{fig4} (a), the cross-time local fusion module adopts temporal sliding window attention within each view. Guided by the attention mask, tokens attend to those within the same window of size $S$ to capture continuous temporal information. Given a specific view, tokens within a time window centered at $t$ are sequentially input to a GRU to obtain the hidden state $h_t$, which serves as the key and value for self-attention with the central query token $t_t$. After processing all tokens, the module outputs the temporal feature $F^{T}$. Note that view tokens only identify the corresponding view and do not participate in attention computation. This module is formulated as:
	\begin{equation}\small
		\setlength\abovedisplayskip{3pt}
		\setlength\belowdisplayskip{3pt}
		\mathit{f}_{CT}([t^I,t^T]) = F^T.
	\end{equation}
	
	These two modules respectively learn spatial representation across views and temporal representations across time steps, significantly enhancing the feature discriminability.
	
	\begin{table*}[t]
		\begin{minipage}[h]{0.5\textwidth}\scriptsize
			\centering
			\setlength{\abovecaptionskip}{0.0cm} 
			\setlength{\belowcaptionskip}{0.2cm} 
			\setlength{\tabcolsep}{1.2pt}
			\centering
			\renewcommand{\arraystretch}{0.9}
			\captionsetup{font={footnotesize}}
			\captionof{table}{Quantitative comparison on camera pose estimation.}
			\addvbuffer[0pt -4pt]{
				\begin{tabular}{ccccccccccc}
					\toprule
					\multirow{2}{*}{Category} & \multirow{2}{*}{Method} & \multicolumn{3}{c}{\tiny Sintel (dynamic) \cite{butler2012naturalistic}} & \multicolumn{3}{c}{\tiny TUM (dynamic) \cite{sturm2012benchmark}} & \multicolumn{3}{c}{\tiny Bonn (dynamic) \cite{palazzolo2019refusion}} \\
					\cmidrule(lr){3-5} \cmidrule(lr){6-8} \cmidrule(lr){9-11}
					& & \tiny ATE \(\downarrow\) & \tiny RTE \(\downarrow\) & \tiny RRE \(\downarrow\) & \tiny ATE \(\downarrow\) & \tiny RTE \(\downarrow\) & \tiny RRE \(\downarrow\) & \tiny ATE \(\downarrow\) & \tiny RTE \(\downarrow\) & \tiny RRE \(\downarrow\) \\
					\midrule
					\multirow{4}{*}{\tiny \shortstack{Specialized\\models}} 
					& \tiny{DPVO} \cite{teed2023deep} & 0.176 & 0.066 & 1.315 & 0.018 & 0.014 & 0.398 & 0.023 & 0.016 & 0.927 \\
					& \tiny{LEAP-VO} \cite{chen2024leap} & \textbf{0.037} & 0.067 & 1.692 & 0.026 & 0.030 & 2.811 & 0.038 & 0.016 & 0.857 \\
					& \tiny{Robust-CVD} \cite{kopf2021robust} & 0.372 & 0.156 & 3.486 & 0.099 & 0.028 & 2.573 & 0.087 & 0.019 & 0.818 \\
					& \tiny{CasualSAM} \cite{zhang2022structure} & 0.141 & \underline{0.041} & 0.648 & 0.037 & 0.017 & 0.758 & 0.025 & 0.015 & 0.864 \\
					\midrule
					\multirow{3}{*}{\tiny \shortstack{Foundation\\models\\(3D)}}
					& \tiny{DUSt3R} \cite{wang2024dust3r} & 0.423 & 0.247 & 5.751 & 0.081 & 0.019 & 3.541 & 0.029 & 0.022 & 1.328 \\
					& \tiny{MASt3R} \cite{leroy2024grounding} & 0.189 & 0.062 & 1.478 & 0.037 & 0.016 & 0.454 & 0.032 & 0.018 & 1.222 \\
					& \tiny{VGGT} \cite{wang2025vggt} & 0.164 & 0.061 & \underline{0.503} & \underline{0.013} & 0.015 & \underline{0.319} & \underline{0.021} & 0.023 & 0.937 \\
					\midrule
					\multirow{4}{*}{\tiny \shortstack{Foundation\\models\\(4D)}}
					& \tiny{MonST3R} \cite{zhang2024monst3r} & 0.111 & 0.045 & 0.741 & 0.094 & 0.021 & 1.359 & 0.024 & \underline{0.014} & \underline{0.795} \\
					& \tiny{StreamVGGT} \cite{zhuo2025streaming} & 0.209 & 0.060 & 0.551 & 0.041 & 0.017 & 0.432 & 0.029 & 0.030 & 1.171 \\
					& \tiny{POMATO} \cite{zhang2025pomato} & 0.183 & 0.052 & 0.598 & 0.034 & \underline{0.013} & 0.519 & 0.036 & 0.026 & 1.132 \\
					& \tiny{Ours} & \underline{0.089} & \textbf{0.029} & \textbf{0.317} & \textbf{0.011} & \textbf{0.009} & \textbf{0.295} & \textbf{0.017} & \textbf{0.013} & \textbf{0.774} \\
					\bottomrule
				\end{tabular}
			}
			\label{tab2:camera pose}
		\end{minipage}
		\begin{minipage}[h]{0.5\textwidth}\scriptsize
			\centering
			\setlength{\abovecaptionskip}{0.0cm} 
			\setlength{\belowcaptionskip}{0.2cm} 
			\setlength{\tabcolsep}{1.2pt}
			\centering
			\renewcommand{\arraystretch}{0.9}
			\captionsetup{font={footnotesize}}
			\captionof{table}{Quantitative comparison on depth stimation.}
			\addvbuffer[0pt -4pt]{
				\begin{tabular}{cccccccc}
					\toprule
					\multirow{3}{*}{Category} & \multirow{3}{*}{Method} & \multicolumn{2}{c}{\tiny Sintel (dynamic) \cite{butler2012naturalistic}} & \multicolumn{2}{c}{\tiny Bonn (dynamic) \cite{palazzolo2019refusion}} & \multicolumn{2}{c}{\tiny KITTI (dynamic) \cite{geiger2013vision}} \\
					\cmidrule(lr){3-4} \cmidrule(lr){5-6} \cmidrule(lr){7-8}
					& & {\tiny Abs Rel \(\downarrow\)} & {\tiny $\delta\textless1.25$ \(\uparrow\)} & {\tiny Abs Rel \(\downarrow\)} & {\tiny $\delta\textless1.25$ \(\uparrow\)}& {\tiny Abs Rel \(\downarrow\)} & {\tiny $\delta\textless1.25$ \(\uparrow\)} \\
					
					\midrule
					\multirow{4}{*}{\tiny \shortstack{Specialized\\models}} 
					&\tiny  NVDS \cite{wang2023neural} & 0.417 & 47.9 & 0.171 & 77.0 & 0.247 & 59.3 \\
					& \tiny DepthCrafter \cite{hu2025depthcrafter} & \underline{0.286} & \underline{70.5} & 0.078 & 96.8 & 0.108 & 88.5 \\
					& \tiny Robust-CVD \cite{kopf2021robust} & 0.716 & 48.1 & 0.203 & 70.1 & 0.282 & 57.4 \\
					& \tiny CausalSAM \cite{zhang2022structure} & 0.395 & 55.1 & 0.165 & 74.2 & 0.251 & 61.7 \\
					\midrule
					\multirow{3}{*}{\tiny \shortstack{Foundation\\models\\(3D)}}
					& \tiny DUSt3R \cite{wang2024dust3r} & 0.605 & 47.6 & 0.146 & 87.1 & 0.136 & 84.1 \\
					& \tiny MASt3R \cite{leroy2024grounding} & 0.568 & 50.3 & 0.221 & 75.6 & 0.158 & 76.7 \\
					& \tiny VGGT \cite{wang2025vggt} & 0.309 & 66.7 & \underline{0.056} & \underline{97.0} & \underline{0.073} & \underline{94.7} \\
					\midrule
					\multirow{4}{*}{\tiny \shortstack{Foundation\\models\\(4D)}}
					& \tiny MonST3R \cite{zhang2024monst3r} & 0.339 & 59.2 & 0.061 & 96.0 & 0.106 & 89.2 \\
					& \tiny StreamVGGT \cite{zhuo2025streaming} & 0.364 & 66.3 & 0.084 & 89.7 & 0.132 & 85.0 \\
					& \tiny POMATO \cite{zhang2025pomato} & 0.351 & 58.6 & 0.074 & 96.1 & 0.082 & 93.0 \\
					& \tiny Ours & \textbf{0.267} & \textbf{72.9} & \textbf{0.048} & \textbf{97.3} & \textbf{0.062} & \textbf{96.8} \\
					\bottomrule
				\end{tabular}
			}
			\label{tab3:depth}
		\end{minipage}
	\end{table*}
	
	\begin{figure*}[t]
		\centering
		\setlength{\abovecaptionskip}{0.1cm} 
		\setlength{\belowcaptionskip}{-0.5cm} 
		\includegraphics[scale=0.12]{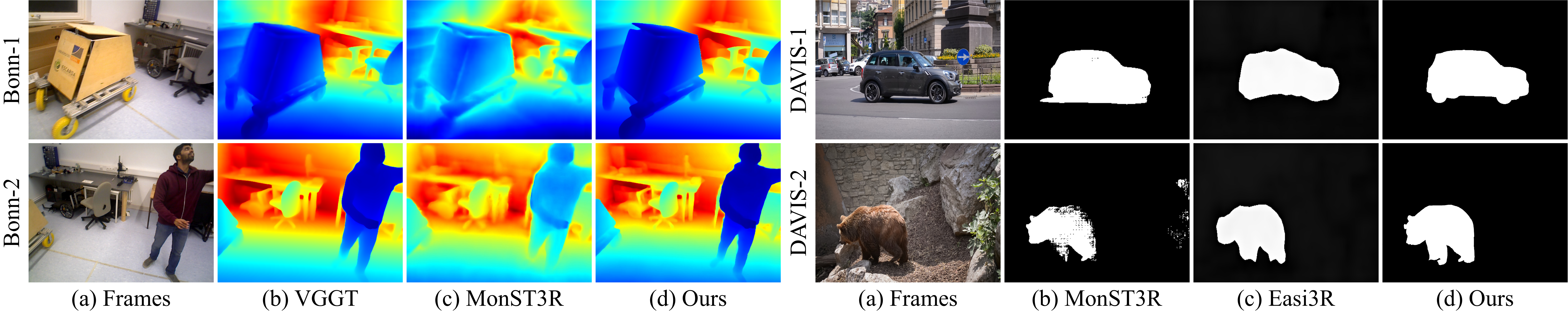}
		\caption{Visual results of depth and dynamic mask estimation.}
		\label{fig5}
	\end{figure*}
	
	\subsection{Multi-task Joint Geometry Prediction}
	\noindent
	\textbf{Task-Specific Heads.} We design five different prediction heads for five geometry tasks: camera parameters, depth, dynamic mask, point map, and tracking estimation. For camera parameters, we propose the spatial head including self-attention layers and linear layers, which are used to derive camera intrinsics and extrinsics $\hat{g}_{vt}$ from the spatial feature $F^S$. For depth map, dynamic mask and point map, we first convert spatial feature $F^S$ and temporal feature $F^T$ into dense feature maps $F^D$ with a DPT module \cite{ranftl2021vision} and apply three different prediction heads to respectively derive depth map $\hat{D}_{vt}$, dynamic mask $\hat{M}_{vt}$ and point map $\hat{P}_{vt}$. For tracking, we follow the design of TAPIR \cite{doersch2023tapir} and CoTracker3 \cite{karaev2024cotracker3}. We initialize $N$ query points $(q_i)_{i=1}^N$ by randomly sampling from the first frame(or any other frame you want). Subsequently, the temporal head utilizes the temporal feature $F^{T}$ as input to derive the 2D trajectory $\hat{T}_{vt}^{2D}$ and 3D trajectory $\hat{T}_{vt}^{3D}$ corresponding to the query points $q_i$ in a coarse-to-fine manner \cite{doersch2023tapir}.
	
	\noindent
	\textbf{Multi-Task Joint Optimization.} For each geometry estimation task, we design a specific loss function. To supervise camera parameter estimation, we apply the Huber loss $\rho_\delta(\cdot)$ between the prediction $\hat{g}_{vt}$ and the ground truth ${g}_{vt}$:
	\begin{equation}\small
		\setlength\abovedisplayskip{3pt}
		\setlength\belowdisplayskip{3pt}
		\mathcal{L}_{cam}=\sum\nolimits_{vt}{\rho_\delta(\hat{g}_{vt}-{g}_{vt})}.
	\end{equation}
	
	Depth estimation is supervised by the ground truth ${D}_{vt}$ with L2 loss and gradient consistency loss:
	\begin{equation}\small
		\setlength\abovedisplayskip{3pt}
		\setlength\belowdisplayskip{3pt}
		\begin{aligned}
			\mathcal{L}_{depth}=\sum\nolimits_{vt}(\parallel \hat{D}_{vt}-{D}_{vt} \parallel_{2}^{2} + \parallel \nabla\hat{D}_{vt}-\nabla{D}_{vt} \parallel).
		\end{aligned}
	\end{equation}
	
	Dynamic mask estimation is supervised by the ground truth ${M}_{vt}$ with binary cross-entropy loss:
	\begin{equation}\small
		\setlength\abovedisplayskip{3pt}
		\setlength\belowdisplayskip{3pt}
		\begin{aligned}
			\mathcal{L}_{mask}&=-\frac{1}{N}\sum\nolimits_{vt}({M}_{vt}\log\hat{M}_{vt}\\
			&+(1-{M}_{vt})\log(1-\hat{M}_{vt})),
		\end{aligned}
	\end{equation}
	where the ground truth ${M}_{vt}$ is binarized and the prediction $\hat{M}_{vt}$ is a probability value in the range of $[0,1]$. 
	
	Point map estimation is supervised by the ground truth ${P}_{vt}$ with L2 loss and gradient consistency loss:
	\begin{equation}\small
		\setlength\abovedisplayskip{3pt}
		\setlength\belowdisplayskip{3pt}
		\begin{aligned}
			\mathcal{L}_{point}=\sum\nolimits_{vt}(\parallel \hat{P}_{vt}-{P}_{vt} \parallel_{2}^{2} + \parallel \nabla\hat{P}_{vt}-\nabla{P}_{vt} \parallel).
		\end{aligned}
	\end{equation}

	For tracking prediction, we calculate Chamfer Distance \cite{fan2017point} between the prediction $\hat{T}_{vt}$ and ground truth ${T}_{vt}$ as the tracking loss, which is formulated as:
	\begin{equation}\small
		\setlength\abovedisplayskip{3pt}
		\setlength\belowdisplayskip{3pt}
		\mathcal{L}_{track}=CD({\hat{T}_{vt}^{2D},{T}_{vt}^{2D}})+CD({\hat{T}_{vt}^{3D},{T}_{vt}^{3D}}).
	\end{equation}
	
	\begin{table*}\footnotesize
		\setlength{\abovecaptionskip}{0.1cm} 
		\setlength{\belowcaptionskip}{0.0cm} 
		\setlength{\tabcolsep}{1.8pt}
		\centering
		\renewcommand{\arraystretch}{0.9}
		\caption{Quantitative comparison on point map estimation.}
		\begin{tabular}{cccccccccccccccccccc}
			\toprule
			\multirow{4}{*}{Category} & \multirow{4}{*}{Method} & \multicolumn{6}{c}{7-Scenes (static) \cite{shotton2013scene}} & \multicolumn{6}{c}{NRGBD (static) \cite{azinovic2022neural}} & \multicolumn{6}{c}{ETH3D (static) \cite{schops2017multi}} \\
			\cmidrule(lr){3-8} \cmidrule(lr){9-14} \cmidrule(lr){15-20}
			& & \multicolumn{2}{c}{Acc. \(\downarrow\)} & \multicolumn{2}{c}{Comp. \(\downarrow\)} & \multicolumn{2}{c}{NC. \(\uparrow\)} & \multicolumn{2}{c}{Acc. \(\downarrow\)} & \multicolumn{2}{c}{Comp. \(\downarrow\)} & \multicolumn{2}{c}{NC. \(\uparrow\)} & \multicolumn{2}{c}{Acc. \(\downarrow\)} & \multicolumn{2}{c}{Comp. \(\downarrow\)} & \multicolumn{2}{c}{NC. \(\uparrow\)}\\
			\cmidrule(lr){3-4} \cmidrule(lr){5-6} \cmidrule(lr){7-8} \cmidrule(lr){9-10} \cmidrule(lr){11-12} \cmidrule(lr){13-14} \cmidrule(lr){15-16} \cmidrule(lr){17-18} \cmidrule(lr){19-20}
			
			& & Mean & Med. & Mean & Med. & Mean & Med. & Mean & Med. & Mean & Med. & Mean & Med.  & Mean & Med. & Mean & Med. & Mean & Med.\\
			
			\midrule
			\multirow{4}{*}{3D} 
			& DUSt3R \cite{wang2024dust3r} & 0.147 & 0.078 & 0.179 & 0.068 & 0.739 & 0.836 & 0.145 & 0.019 & 0.155 & 0.018 & 0.871 & 0.981 & 0.748 & 0.617 & 0.789 & 0.676 & 0.691 & 0.813\\
			& Spann3R \cite{wang20243d} & 0.295 & 0.224 & 0.206 & 0.113 & 0.651 & 0.728 & 0.414 & 0.321 & 0.415 & 0.283 & 0.683 & 0.787 & 0.465 & 0.372 & 0.472 & 0.364 & 0.731 & 0.825\\
			& VGGT \cite{wang2025vggt} & \underline{0.087} & \textbf{0.039} & \underline{0.092} & 0.039 & \underline{0.788} & \underline{0.889} & 0.072 & 0.018 & 0.078 & 0.021 & 0.911 & \underline{0.989} & 0.281 & 0.186 & 0.306 & 0.183 & 0.852 & 0.949\\
			& $\pi$3 \cite{wang2025pi} & 0.095 & 0.049 & 0.108 & 0.052 & 0.776 & 0.880 & \underline{0.055} & \underline{0.017} & \textbf{0.049} & \underline{0.017} & \underline{0.915} & 0.974 & \underline{0.195} & \underline{0.132} & \underline{0.211} & \underline{0.128} & \textbf{0.884} & \underline{0.968}\\
			
			\midrule
			\multirow{5}{*}{4D} 
			& MonST3R \cite{zhang2024monst3r} & 0.249 & 0.186 & 0.268 & 0.168 & 0.673 & 0.758 & 0.273 & 0.115 & 0.288 & 0.111 & 0.759 & 0.842 & 0.492 & 0.386 & 0.503 & 0.389 & 0.745 & 0.818\\
			& CUT3R \cite{wang2025continuous} & 0.127 & 0.051 & 0.155 & \underline{0.031} & 0.728 & 0.832 & 0.098 & 0.031 & 0.076 & 0.027 & 0.838 & 0.970 & 0.618 & 0.526 & 0.748 & 0.580 & 0.755 & 0.847\\
			& StreamVGGT \cite{zhuo2025streaming} & 0.130 & 0.056 & 0.116 & 0.042 & 0.752 & 0.864 & 0.085 & 0.044 & 0.075 & 0.041 & 0.862 & 0.985 & 0.328 & 0.252 & 0.337 & 0.244 & 0.815 & 0.912\\
			& Mem4D \cite{cai2025mem4d} & 0.184 & 0.136 & 0.177 & 0.082 & 0.740 & 0.814 & 0.270 & 0.195 & 0.213 & 0.114 & 0.734 & 0.823 & 0.478 & 0.365 & 0.491 & 0.359 & 0.738 & 0.809\\
			& Ours & \textbf{0.082} & \underline{0.048} & \textbf{0.073} & \textbf{0.030} & \textbf{0.896} & \textbf{0.941} & \textbf{0.050} & \textbf{0.016} & \underline{0.052} & \textbf{0.015} & \textbf{0.925} & \textbf{0.992} & \textbf{0.172} & \textbf{0.120} & \textbf{0.187} & \textbf{0.118} & \underline{0.879} & \textbf{0.972}\\
			\bottomrule
		\end{tabular}
		\label{tab4:point map}
	\end{table*}
	
	\begin{figure*}[t]
		\centering
		\setlength{\abovecaptionskip}{0.3cm} 
		\setlength{\belowcaptionskip}{-0.5cm} 
		\includegraphics[scale=0.1]{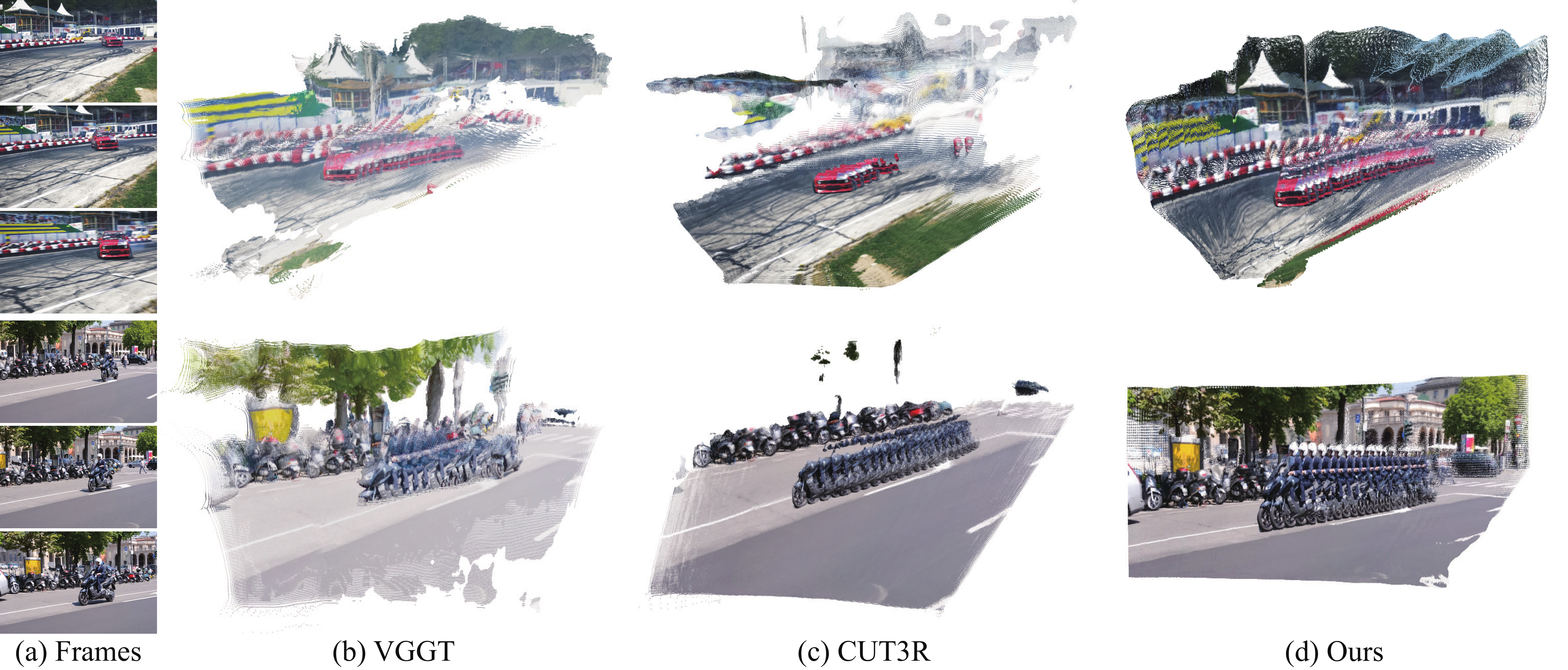}
		\caption{Visual results of point map estimation.}
		\label{fig6}
	\end{figure*}
	
	Overall, we train our 4D-VGGT model with the multi-task loss, which is formulated as:
	\begin{equation}\small
		\setlength\abovedisplayskip{3pt}
		\setlength\belowdisplayskip{3pt}
		\label{equa:loss}
		\begin{aligned}
			\mathcal{L}&=\lambda_{cam} \mathcal{L}_{cam}+\lambda_{depth}\mathcal{L}_{depth}+\lambda_{mask}\mathcal{L}_{mask} \\
			&+\lambda_{point}\mathcal{L}_{point}+\lambda_{track}\mathcal{L}_{track},
		\end{aligned}
	\end{equation}
	where $\lambda_{cam}$, $\lambda_{depth}$, $\lambda_{mask}$, $\lambda_{point}$, $\lambda_{track}$ are the weights that control the importance of the related losses.

	\section{Training Datasets and Pipeline}
	
	\subsection{Training Datasets}
	We train our 4D-VGGT model on a diverse collection of datasets, which together provide a wide range of annotations including camera pose, depth map, dynamic mask, point cloud and trajectory. The training datasets we used are shown in the Tab. \ref{tab1:datasets}. For camera pose, we use datasets containing a total of 7.1 million samples, \emph{e.g.}, Co3Dv2 \cite{reizenstein2021common} and WildRGB-D \cite{xia2024rgbd}. For depth, we collect datasets including 15.6 million samples, \emph{e.g.}, BlendedMVS \cite{yao2020blendedmvs} and DL3DV \cite{ling2024dl3dv}. For dynamic mask, point map, and tracking, we select datasets containing 5.8 million, 1.6 million, and 1.2 million samples, respectively, including Kubric \cite{greff2022kubric}, MegaDepth \cite{li2018megadepth}, TAP-Vid \cite{doersch2022tap}, \emph{etc}. For joint optimization, we use datasets containing a total of 8.2 million samples, \emph{e.g.}, TartanAir \cite{wang2020tartanair} and Spring \cite{mehl2023spring}.
	
	\subsection{Training Pipeline}
	As shown in Tab. \ref{tab1:datasets}, our model is trained in two stages: per-task optimization and joint optimization.
	
	\noindent
	\textbf{Stage 1: Per-Task Optimization.} This stage equips the model with spatiotemporal representation and geometry estimation capabilities. We optimize the DINO encoder, feature representation modules, and the prediction heads for each geometry task using their task-specific losses. Specifically, we train the model using camera pose, depth, dynamic mask, point map and tracking tasks sequentially. Note that when training on one task, other prediction heads are frozen.
	
	\noindent
	\textbf{Stage 2: Multi-Task Optimization.} This stage enhances the performance of multi-task joint prediction. We freeze the encoder and feature representation modules and fine-tune all prediction heads using the multi-task joint loss.
	
	\begin{table*}[t]
		\begin{minipage}[h]{0.5\textwidth}\scriptsize
			\centering
			\setlength{\abovecaptionskip}{0.0cm} 
			\setlength{\belowcaptionskip}{0.0cm} 
			\setlength{\tabcolsep}{2.4pt}
			\centering
			\renewcommand{\arraystretch}{0.9}
			\captionsetup{font={footnotesize}}
			\captionof{table}{Quantitative comparison on dynamic mask estimation.}
			\addvbuffer[0pt -12pt]{
				\begin{tabular}{cccccccc}
					\toprule
					\multirow{3}{*}{Category} & \multirow{3}{*}{Method} & \multicolumn{2}{c}{DAVIS-16 \cite{perazzi2016benchmark}} & \multicolumn{2}{c}{DAVIS-17 \cite{pont20172017}} & \multicolumn{2}{c}{SegTrackv2 \cite{li2013video}} \\
					\cmidrule(lr){3-4} \cmidrule(lr){5-6} \cmidrule(lr){7-8}
					& & {J\textsubscript{M} \(\uparrow\)} &{J\textsubscript{R} \(\uparrow\)} & {J\textsubscript{M} \(\uparrow\)} &{J\textsubscript{R} \(\uparrow\)}& {J\textsubscript{M} \(\uparrow\)} &{J\textsubscript{R} \(\uparrow\)} \\
					
					\midrule
					\multirow{2}{*}{\shortstack{Specialized\\models}} & OCLR \cite{xie2022segmenting} & {67.4} & {78.2} & {62.1} & {69.8} & \underline{64.7} & 68.5 \\
					& ABR \cite{xie2024appearance} & \underline{70.9} & \underline{81.3} & \underline{64.8} & \underline{73.5} & \textbf{65.1} & \textbf{72.9} \\
					
					\midrule
					\multirow{5}{*}{\shortstack{Foundation\\models}} & D$^2$USt3R \cite{han2025d} & 50.7 & 56.1 & 44.3 & 45.5 & 41.2 & 41.3 \\
					& MonST3R \cite{zhang2024monst3r} & 41.2 & 46.1 & 35.6 & 35.7 & 33.1 & 31.8 \\
					& Easi3R \cite{chen2025easi3r} & 57.4 & 71.2 & 56.8 & 68.3 & 52.1 & 58.3 \\
					& POMATO \cite{zhang2025pomato} & 49.5 & 55.2 & 43.4 & 44.3 & 43.1 & 43.9 \\
					& Ours & \textbf{71.0} & \textbf{82.2} & \textbf{67.1} & \textbf{75.0} & 64.5 & \underline{70.6} \\
					
					\bottomrule
				\end{tabular}
			}
			\label{tab5:mask}
		\end{minipage}
		\begin{minipage}[h]{0.5\textwidth}\scriptsize
			\centering
			\setlength{\abovecaptionskip}{0.0cm} 
			\setlength{\belowcaptionskip}{0.0cm} 
			\setlength{\tabcolsep}{1.2pt}
			\centering
			\renewcommand{\arraystretch}{0.9}
			\captionsetup{font={footnotesize}}
			\captionof{table}{Quantitative comparison on point tracking estimation.}
			\addvbuffer[0pt -12pt]{
				\begin{tabular}{cccccccc}
					\toprule
					\multirow{3}{*}{Category} & \multirow{3}{*}{Method} & \multicolumn{2}{c}{PointOdyssey \cite{zheng2023pointodyssey}} & \multicolumn{2}{c}{ADT \cite{pan2023aria}} & \multicolumn{2}{c}{PStudio \cite{joo2015panoptic}} \\
					\cmidrule(lr){3-4} \cmidrule(lr){5-6} \cmidrule(lr){7-8}
					& & {L-12 \(\uparrow\)} &{L-24 \(\uparrow\)} & {L-12 \(\uparrow\)} &{L-24 \(\uparrow\)} &{L-12 \(\uparrow\)} &{L-24 \(\uparrow\)} \\
					
					\midrule
					\multirow{3}{*}{\shortstack{Specialized\\models}} 
					& TAPIR \cite{doersch2023tapir} & 18.31 & 17.85 & 20.01 & 19.11 & 24.39 & 22.65 \\
					& SpatialTracker \cite{xiao2024spatialtracker} & 21.52 & 20.63 & 21.59 & 20.61 & 26.37 & 23.81  \\
					& CoTracker3 \cite{karaev2024cotracker3} & 24.79 & 23.89 & 27.53 & 25.57 & \underline{27.85} & \underline{24.10} \\
					\midrule
					\multirow{4}{*}{\shortstack{Foundation\\models}} 
					& MonST3R \cite{zhang2024monst3r} & 27.35 & 27.88 & 28.27 & 26.09 & 16.48 & 11.02 \\
					& St4RTrack \cite{feng2025st4rtrack} & \underline{35.72} & \underline{34.45} & \underline{34.89} & \underline{31.82} & {26.41} & {20.78} \\
					& POMATO \cite{zhang2025pomato} & 33.18 & 33.54 & 31.61 & 28.26 & 24.63 & 19.83 \\
					& \textbf{Ours} & \textbf{36.12} & \textbf{34.98} & \textbf{35.27} & \textbf{32.11} & \textbf{28.79} & \textbf{25.03} \\
					\bottomrule
				\end{tabular}
			}
			\label{tab6:tracking}
		\end{minipage}
	\end{table*}
	
	\subsection{Implementation Details}
	Specifically, the size $V\times T$ of the view-time grid in the adaptive visual grid is determined by the number of views and time steps of the input sequence. Regarding the attention mask, during inference, we directly generate two masks based on the input sequence settings to guide the subsequent two modules. During training, we perform random sampling based on the input sequence settings to simulate more possible settings. In the cross-view global fusion module, we repeat the intra-frame attention and the inter-frame attention for $L=16$ times. In the cross-time local fusion module, we employ a sliding window of size $S=5$ by default. For model training, we optimize the loss function as shown in Eq. (\ref{equa:loss}) using the AdamW optimizer with an initial learning rate of $1e-5$ and a weight decay of $0.01$. For the weights in Eq. (\ref{equa:loss}), we set $\lambda_{cam}=1.0, \lambda_{depth}=0.8, \lambda_{mask}=0.8, \lambda_{point}=0.9, \lambda_{track}=0.1$. The training and evaluating processes are completed on 8 NVIDIA L20 GPUs.
	
	\section{Experiments}
	\label{sec:experiments}
	
	\subsection{Comparison with State-of-the-Art Models}
	\noindent
	\textbf{Comparison on Camera Pose.} We evaluate our model for camera pose estimation on the Sintel \cite{butler2012naturalistic}, TUM-dynamics \cite{sturm2012benchmark}, and Bonn \cite{palazzolo2019refusion} datasets. Following LEAP-VO \cite{chen2024leap}, we use the same evaluation split for Sintel, while all samples from TUM-dynamics and Bonn are included. Evaluation metrics include Absolute Translation Error (ATE), Relative Translation Error (RTE), and Relative Rotation Error (RRE). We compare our model with specialized pose estimation methods, including DPVO \cite{teed2023deep} and LEAP-VO, Robust-CVD \cite{kopf2021robust}, and CausalSAM \cite{zhang2022structure}. Moreover, we compare against foundation models for static scene geometry, such as DUSt3R \cite{wang2024dust3r}, MASt3R \cite{leroy2024grounding}, and VGGT \cite{wang2025vggt}, as well as those for dynamic scene geometry, including MonST3R \cite{zhang2024monst3r}, StreamVGGT \cite{zhuo2025streaming}, and POMATO \cite{zhang2025pomato}.
	
	As shown in Tab. \ref{tab2:camera pose}, 4D-VGGT achieves consistently strong performance across all datasets and metrics, outperforming existing specialized models and foundation models for both static and dynamic scene geometry.
	
	\begin{figure}[t]
		\centering
		\setlength{\abovecaptionskip}{0.1cm} 
		\setlength{\belowcaptionskip}{-0.7cm} 
		\includegraphics[scale=0.165]{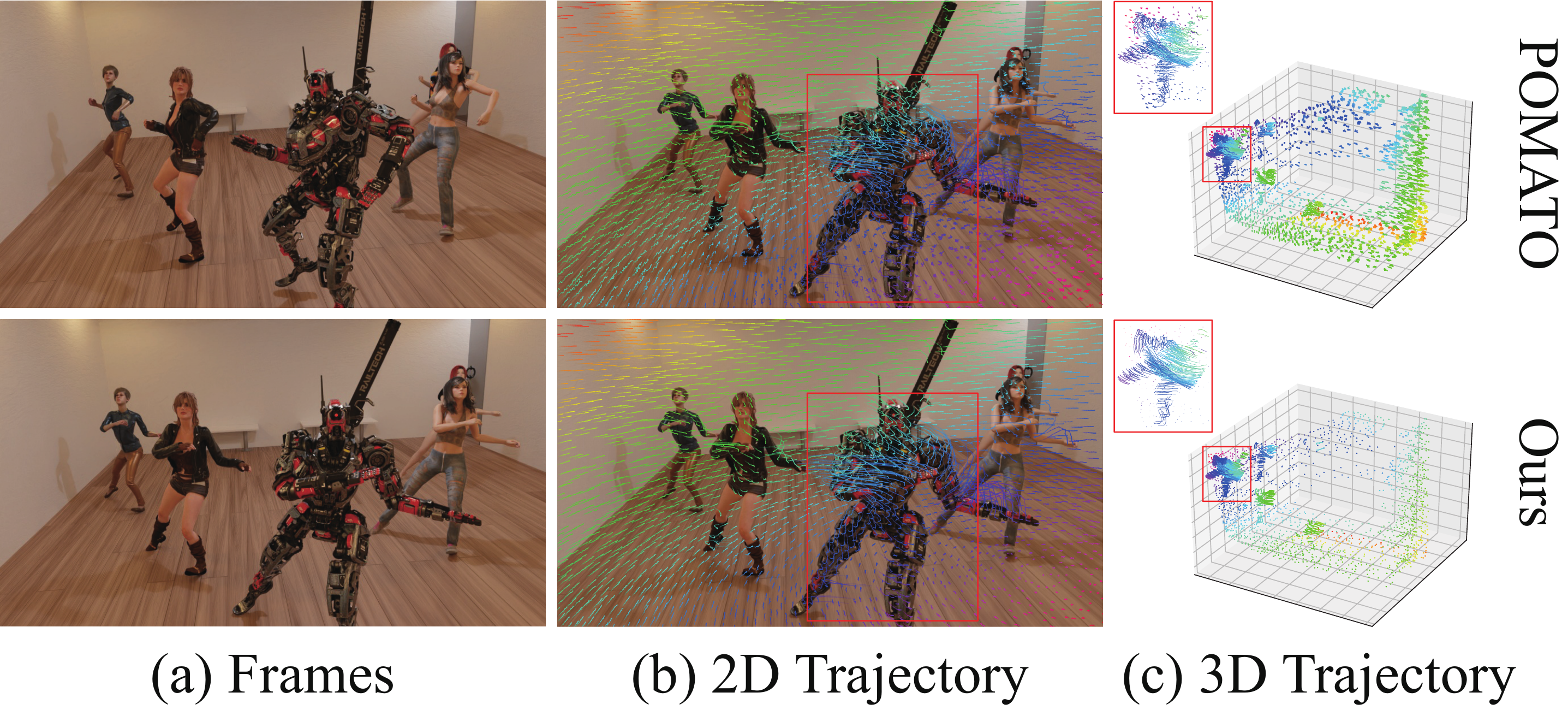}
		\caption{Visual results of tracking.}
		\label{fig7}
	\end{figure}
	
	\noindent
	\textbf{Comparison on Depth.} Following DepthCrafter \cite{hu2025depthcrafter}, we evaluate depth estimation on the Sintel, Bonn, and KITTI \cite{geiger2013vision} datasets. We apply per-scene alignment to ensure scale- and shift-invariant comparisons. For qualitative evaluation, we visualize depth maps on the Bonn dataset. For quantitative evaluation, we report the absolute relative error (Abs Rel) and the inlier ratio ($\delta\textless1.25$). Our model is compared with specialized methods (NVDS \cite{wang2023neural}, DepthCrafter, Robust-CVD, CausalSAM) and foundation models following the same protocol as in the camera pose evaluation.
	
	As shown in Fig. \ref{fig5} and Tab. \ref{tab3:depth}, our method achieves superior qualitative and quantitative results. It estimates accurate global structure and fine local details, and outperforms both specialized and foundation models across all datasets, demonstrating its strong depth estimation capability.
	
	\noindent
	\textbf{Comparison on Dynamic Mask.} We evaluate dynamic mask estimation on the DAVIS-16 \cite{perazzi2016benchmark}, DAVIS-17 \cite{pont20172017}, and SegTrackv2 \cite{li2013video} datasets. Following DAVIS \cite{perazzi2016benchmark}, we adopt the mean IoU (J\textsubscript{M}) and the recall of IoU (J\textsubscript{R}) as evaluation metrics. We compare our model with specialized models including OCLR \cite{xie2022segmenting} and ABR \cite{xie2024appearance}, as well as foundation models for dynamic scene geometry, including MonST3R, D$^2$USt3R \cite{han2025d}, Easi3R \cite{chen2025easi3r}, and POMATO.
	
	As shown in Fig. \ref{fig5} and Tab. \ref{tab5:mask}, our 4D-VGGT surpasses all foundation models and achieves competitive performance with specialized segmentation methods.
	
	\noindent
	\textbf{Comparison on Point Map.} We evaluate point map accuracy on the 7-Scenes \cite{shotton2013scene}, NRGBD \cite{azinovic2022neural}, and ETH3D \cite{schops2017multi} datasets. Following VGGT and CUT3R \cite{wang2025continuous}, we use sparsely sampled inputs: 3–5 frames per scene for 7-Scenes, 2–4 for NRGBD, and 10 for ETH3D. Quantitative metrics include Accuracy (Acc.), Completion (Comp.), and Normal Consistency (NC.), with point maps aligned to ground truth via the Umeyama algorithm \cite{umeyama2002least}. Our model is compared against foundation models for static scene geometry, including DUSt3R, Spann3R \cite{wang20243d}, VGGT, and $\pi$3 \cite{wang2025pi}, as well as those for dynamic scene geometry, including MonST3R, CUT3R, StreamVGGT, and Mem4D \cite{cai2025mem4d}.
	
	As illustrated in Fig. \ref{fig6} and Tab. \ref{tab4:point map}, 4D-VGGT delivers visually superior point maps and achieves state-of-the-art performance across most metrics, consistently outperforming all static and dynamic geometry estimation methods.
	
	\begin{table} \scriptsize
		\setlength{\abovecaptionskip}{0.0cm} 
		\setlength{\belowcaptionskip}{0.0cm} 
		\setlength{\tabcolsep}{0.9pt}
		\centering
		\renewcommand{\arraystretch}{0.8}
		\caption{Effectiveness of our proposed modules.}
		\addvbuffer[0pt -8pt]{
			\begin{tabular}{ccccccccccc}
				\toprule
				\multicolumn{3}{c}{Modules} & \multicolumn{3}{c}{Camera pose} & \multicolumn{3}{c}{Point map}  & \multicolumn{2}{c}{Tracking}  \\
				\cmidrule(lr){1-3} \cmidrule(lr){4-6} \cmidrule(lr){7-9} \cmidrule(lr){10-11}
				{AVG} & {CVGF} & {CTLF} & ATE \(\downarrow\) & RTE \(\downarrow\) & RRE \(\downarrow\) & Acc. \(\downarrow\) & Comp. \(\downarrow\) & NC. \(\uparrow\) & L-12 \(\uparrow\) & L-24 \(\uparrow\)\\
				\midrule
				\xmark & \xmark & \xmark & 0.045 & 0.038 & 2.095 & 0.291 & 0.329 & 0.714 & 17.36 & 16.41\\
				\xmark & \cmark & \xmark & 0.027 & 0.022 & 1.318 & 0.209 & 0.224 & 0.793 & 17.36 & 16.41\\
				\xmark & \xmark & \cmark & 0.045 & 0.038 & 2.095 & 0.234 & 0.256 & 0.762 & 25.81 & 23.68\\
				\xmark & \cmark & \cmark & 0.027 & 0.022 & 1.318 & 0.177 & 0.183 & 0.836 & 25.81 & 23.68\\
				\cmark & \xmark & \xmark & 0.030 & 0.026 & 1.492 & 0.215 & 0.237 & 0.780 & 24.76 & 23.43\\
				\cmark & \cmark & \xmark & \textbf{0.017} & \textbf{0.013} & \textbf{0.774} & 0.085 & 0.088 & 0.895 & 24.76 & 23.43\\
				\cmark & \xmark & \cmark & 0.030 & 0.026 & 1.492 & 0.091 & 0.094 & 0.883 & \textbf{36.12} & \textbf{34.98}\\
				\cmark & \cmark & \cmark & \textbf{0.017} & \textbf{0.013} & \textbf{0.774} & \textbf{0.050} & \textbf{0.052} & \textbf{0.925} & \textbf{36.12} & \textbf{34.98}\\
				\bottomrule
			\end{tabular}
		}
		\label{tab7:ablation on module}
	\end{table}
	
	\noindent
	\textbf{Comparison on Tracking.} We evaluate the tracking performance on PointOdyssey \cite{zheng2023pointodyssey}, Aria Digital Twin (ADT) \cite{pan2023aria}, and Panoptic Studio (PStudio) \cite{joo2015panoptic}. Following prior work, we report Average Percent Deviation (APD) for trajectories of 12 and 24 frames (L-12 and L-24). We compare our model with specialized models such as TAPIR \cite{doersch2023tapir}, SpatialTracker \cite{xiao2024spatialtracker}, and CoTracker3 \cite{karaev2024cotracker3}, as well as foundation models, \emph{e.g.}, MonST3R, St4RTrack \cite{feng2025st4rtrack}, and POMATO.
	
	As illustrated in Fig. \ref{fig7} and Tab. \ref{tab6:tracking}, our 4D-VGGT produces smooth, coherent 2D and 3D trajectories and consistently surpasses all specialized and foundation models.
	
	\subsection{Ablation Study}
	\noindent
	\textbf{How Proposed Modules Work.} We conduct ablation studies on camera pose, point map, and point tracking to evaluate the impact of the Adaptive Visual Grid (AVG), Cross-View Global Fusion (CVGF), and Cross-Time Local Fusion (CTLF) modules. As shown in Tab. \ref{tab7:ablation on module}, both modules significantly improve the performance of point map estimation, while CVGF and CTLF further enhance camera pose estimation and point tracking, respectively.
	
	\noindent
	\textbf{Influence of Training Strategy.} We further conduct ablation experiments to verify the effectiveness of our training strategy. Our model adopts a two-stage (TS) manner and a random sampling strategy. We compare our training paradigm with those that employ single-stage (SS) training or without the random sampling strategy. As shown in Tab. \ref{tab8:ablation on training strategy}, both components of our training strategy lead to notable performance gains across all tasks.
	
	\begin{table} \scriptsize
		\setlength{\abovecaptionskip}{0.0cm} 
		\setlength{\belowcaptionskip}{0.0cm} 
		\setlength{\tabcolsep}{1.3pt}
		\centering
		\renewcommand{\arraystretch}{0.8}
		\caption{Influence of training strategy.}
		\addvbuffer[0pt -8pt]{
			\begin{tabular}{ccccccccc}
				\toprule
				\multirow{3}{*}{Training Strategy} & \multicolumn{3}{c}{Camera pose} & \multicolumn{3}{c}{Point map}  & \multicolumn{2}{c}{Tracking}  \\
				\cmidrule(lr){2-4} \cmidrule(lr){5-7} \cmidrule(lr){8-9}
				& ATE \(\downarrow\) & RTE \(\downarrow\) & RRE \(\downarrow\) & Acc. \(\downarrow\) & Comp. \(\downarrow\) & NC. \(\uparrow\) & L-12 \(\uparrow\) & L-24 \(\uparrow\)\\
				\midrule
				w/SS & 0.026 & 0.024 & 1.291 & 0.093 & 0.087 & 0.853 & 26.61 & 24.93\\
				w/SS+Sampling & 0.024 & 0.021 & 1.048 & 0.081 & 0.083 & 0.877 & 28.59 & 25.86\\
				w/TS & 0.020 & 0.015 & 0.854 & 0.065 & 0.066 & 0.905 & 33.49 & 32.07\\
				\textbf{w/TS+Sampling} & \textbf{0.017} & \textbf{0.013} & \textbf{0.774} & \textbf{0.050} & \textbf{0.052} & \textbf{0.925} & \textbf{36.12} & \textbf{34.98} \\
				\bottomrule
			\end{tabular}
		}
		\label{tab8:ablation on training strategy}
	\end{table}
	
	\begin{table} \scriptsize
		\setlength{\abovecaptionskip}{0.0cm} 
		\setlength{\belowcaptionskip}{0.0cm} 
		\setlength{\tabcolsep}{1.3pt}
		\centering
		\renewcommand{\arraystretch}{0.8}
		\caption{Impact of spatial and temporal attention masks.}
		\addvbuffer[0pt -8pt]{
			\begin{tabular}{ccccccccc}
				\toprule
				\multirow{3}{*}{Attention Mask} & \multicolumn{3}{c}{Camera pose} & \multicolumn{3}{c}{Point map}  & \multicolumn{2}{c}{Tracking}  \\
				\cmidrule(lr){2-4} \cmidrule(lr){5-7} \cmidrule(lr){8-9}
				& ATE \(\downarrow\) & RTE \(\downarrow\) & RRE \(\downarrow\) & Acc. \(\downarrow\) & Comp. \(\downarrow\) & NC. \(\uparrow\) & L-12 \(\uparrow\) & L-24 \(\uparrow\)\\
				\midrule
				w/o & 0.024 & 0.022 & 1.127 & 0.133 & 0.141 & 0.882 & 29.53 & 28.41\\
				w/Spatial-mask & \textbf{0.017} & \textbf{0.013} & \textbf{0.774} & 0.075 & 0.077 & 0.901 & 29.53 & 28.41\\
				w/Temporal-mask & 0.024 & 0.022 & 1.127 & 0.082 & 0.089 & 0.892 & \textbf{36.12} & \textbf{34.98}\\
				\textbf{w/Both} & \textbf{0.017} & \textbf{0.013} & \textbf{0.774} & \textbf{0.050} & \textbf{0.052} & \textbf{0.925} & \textbf{36.12} & \textbf{34.98}\\
				\bottomrule
			\end{tabular}
		}
		\label{tab9:analysis1}
	\end{table}
	
	\begin{figure}[t]
		\centering
		\setlength{\abovecaptionskip}{0.3cm} 
		\setlength{\belowcaptionskip}{-0.5cm} 
		\includegraphics[scale=0.39]{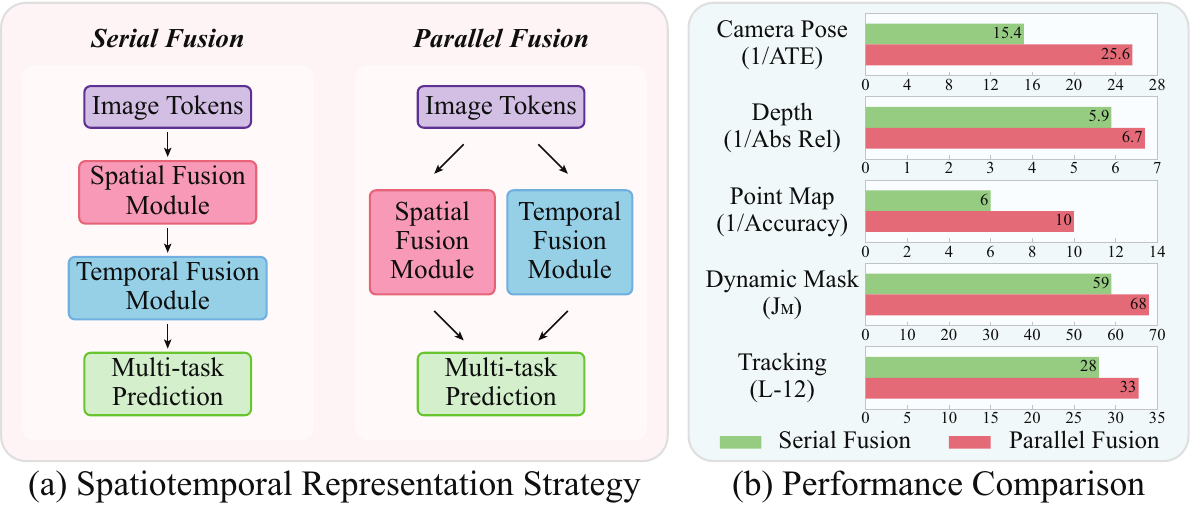}
		\caption{\textbf{Analysis on spatiotemporal representation strategy.} To validate the effectiveness of our parallel fusion strategy, we compare it with serial fusion strategy across multiple geometry estimation tasks. As shown in the bar chart, our strategy consistently yields superior performance in all tasks.}
		\label{fig8}
	\end{figure}
	
	\subsection{Discussion}
	\noindent
	\textbf{Effectiveness of Spatiotemporal Representation Strategy.} A key difference between 4D-VGGT and existing 4D geometry estimation methods lies in the parallel fusion strategy for spatiotemporal representation, while prior methods adopt the serial fusion strategy. To verify its effectiveness, we evaluate the performance across multiple geometry estimation tasks. As shown in Fig. \ref{fig8}, the parallel fusion strategy yields notably better performance, particularly on tasks sensitive to dynamic patterns, \emph{e.g.}, camera pose and point map.
	
	\noindent
	\textbf{Impact of Attention Masks.} We evaluate the effect of attention masks in the spatiotemporal representation module through quantitative experiments on two mask types: the spatial mask and the temporal mask. As shown in Tab. \ref{tab9:analysis1}, both masks improve point map estimation performance, while the spatial and temporal masks notably enhance camera pose estimation and point tracking, respectively.
	
	\noindent
	\textbf{Effect of Hyperparameters.} To determine the optimal hyperparameters for the spatiotemporal representation module, we conduct additional quantitative experiments. As shown in Tab. \ref{tab10:analysis2}, increasing the number of layers $L$ in CVGF leads to performance improvements but higher inference time. Therefore, we set $L=16$ to achieve the best trade-off between accuracy and efficiency. For CTLF, we adopt the same principle and set the window size $S=5$.
	
	\begin{table}\scriptsize
		\setlength{\abovecaptionskip}{0.0cm} 
		\setlength{\belowcaptionskip}{0.0cm} 
		\setlength{\tabcolsep}{0.8pt}
		\centering
		\renewcommand{\arraystretch}{1.0}
		\caption{Effect of hyperparameters.}
		\addvbuffer[0pt -8pt]{
			\begin{tabular}{ccccccccccc}
				\toprule
				\multirow{3}{*}{Parameter} & \multirow{3}{*}{Value} & \multicolumn{3}{c}{Camera pose} & \multicolumn{3}{c}{Point map}  & \multicolumn{2}{c}{Tracking} &  \multirow{3}{*}{\tiny \shortstack{Inference\\Time(s)}}  \\ 
				\cmidrule(lr){3-5} \cmidrule(lr){6-8} \cmidrule(lr){9-10}
				& & ATE\(\downarrow\) & RTE\(\downarrow\) & RRE\(\downarrow\) & Acc.\(\downarrow\) & Comp.\(\downarrow\) & NC.\(\uparrow\) & L-12\(\uparrow\) & L-24\(\uparrow\)& \\
				\midrule
				\multirow{3}{*}{\tiny \shortstack{Layers $L$\\in CVGF }} & 12 & 0.022 & 0.021 & 0.923 & 0.068 & 0.069 & 0.894 & 36.12 & 34.98 & 2.56 \\
				& \textbf{16} & 0.017 & 0.013 & 0.774 & 0.050 & 0.052 & 0.925 & 36.12 & 34.98 & 3.19 \\
				& 20 & 0.016 & 0.014 & 0.768 & 0.048 & 0.049 & 0.929 & 36.12 & 34.98 & 3.85 \\ \hline
				\multirow{4}{*}{\tiny \shortstack{Window size $S$\\in CTLF }} & 3 & 0.017 & 0.013 & 0.774 & 0.071 & 0.074 & 0.908 & 31.52 & 29.83 & 2.06 \\
				& \textbf{5} & 0.017 & 0.013 & 0.774 & 0.050 & 0.052 & 0.925 & 36.12 & 34.98 & 3.19 \\
				& 7 & 0.017 & 0.013 & 0.774 & 0.049 & 0.052 & 0.931 & 36.43 & 35.77 & 4.32 \\
				& 9 & 0.017 & 0.013 & 0.774 & 0.050 & 0.051 & 0.928 & 36.67 & 36.42 & 5.27 \\
				\bottomrule
			\end{tabular}
		}
		\label{tab10:analysis2}
	\end{table}
	
	\begin{table} \scriptsize
		\setlength{\abovecaptionskip}{0.1cm} 
		\setlength{\belowcaptionskip}{0.0cm} 
		\setlength{\tabcolsep}{1.3pt}
		\centering
		\renewcommand{\arraystretch}{0.8}
		\caption{Generalization for camera settings.}
		\addvbuffer[0pt -12pt]{
			\begin{tabular}{ccccccccc}
				\toprule
				\multirow{3}{*}{Camera setting} & \multicolumn{3}{c}{Camera pose} & \multicolumn{3}{c}{Point map}  & \multicolumn{2}{c}{Tracking}  \\
				\cmidrule(lr){2-4} \cmidrule(lr){5-7} \cmidrule(lr){8-9}
				& ATE \(\downarrow\) & RTE \(\downarrow\) & RRE \(\downarrow\) & Acc. \(\downarrow\) & Comp. \(\downarrow\) & NC. \(\uparrow\) & L-12 \(\uparrow\) & L-24 \(\uparrow\)\\
				\midrule
				Mono-S & 0.023 & 0.019 & 0.920 & 0.063 & 0.065 & 0.797 & 30.90 & 29.45 \\
				Mono-D & 0.021 & 0.017 & 0.880 & 0.059 & 0.061 & 0.815 & 31.35 & 30.05\\
				Multi-S & 0.021 & 0.016 & 0.845 & 0.058 & 0.058 & 0.832 & 32.72 & 31.60\\
				\bottomrule
			\end{tabular}
		}
		\label{tab11:analysis3}
	\end{table}
	
	\noindent
	\textbf{Generalization on Camera Settings.} To verify the generalization of our model to different camera settings, we conduct experiments on the same scenes under multi-view static (Multi-S), monocular static (Mono-S), and monocular dynamic (Mono-D) settings. As shown in Tab. \ref{tab11:analysis3}, our model maintains consistent performance across all tasks, demonstrating strong generalization to diverse camera settings.
	
	\noindent
	\textbf{Limitations.} While our model effectively captures dynamic scene geometry from various camera settings, it still faces several challenges. Due to the fixed window size in the CTLF module, the motion patterns of fast-moving objects may not be fully captured. In future work, we aim to integrate event cameras \cite{zhou2025bridge,wang2025injecting} to provide visual signals with higher temporal resolution and better capture rapid dynamics.

	\section{Conclusion}
	
	In this work, we propose 4D-VGGT, a general spatiotemporal foundation model for dynamic scene geometry estimation. We design a multi-setting input component, where we introduce an adaptive visual grid to support inputs from arbitrary views and time steps. We propose a multi-level representation component, which achieves a divide-and-conquer spatiotemporal representation through attention masks. We construct a multi-task prediction component, which builds multiple task-specific heads to learn the dynamic scene geometry via multi-task optimization.
	The three components jointly promote the feature discriminability and application universality for dynamic scenes. We further integrate diverse datasets to train our model and validate the effectiveness of our method through extensive experiments. We believe that this work provides a new paradigm for the research community on multi-task prediction in dynamic scenes.

	{
		\small
		\bibliographystyle{ieeenat_fullname}
		\bibliography{egbib}
	}

\end{document}